# SynthVision - Harnessing Minimal Input for Maximal Output in Computer Vision Models using Synthetic Image data

Yudara Kularathne MD[1*], Prathapa Janitha[1*], Sithira Ambepitiya MBBS[1*]

Thanveer Ahamed[1+], Dinuka Wijesundara[1+], Prarththanan Sothyrajah[1+]

(1) HeHealth Inc. San Francisco, USA

## Abstract

Rapid development of disease detection computer vision models is vital in response to urgent medical crises like epidemics or events of bioterrorism. However, traditional data-gathering methods are too slow for these scenarios, necessitating innovative approaches to generate reliable models quickly from minimal data. We demonstrate our new approach by building a comprehensive computer vision model for detecting Human Papilloma Virus (HPV) Genital warts using only synthetic data. In our study, we employed a two-phase experimental design using diffusion models. In the first phase, diffusion models were utilized to generate a large number of diverse synthetic images from 10 HPV guide images, explicitly focusing on accurately depicting genital warts. The second phase involved the training and testing a vision model using this synthetic dataset. This method aimed to assess the effectiveness of diffusion models in rapidly generating high-quality training data and the subsequent impact on the vision model's performance in medical image recognition. The study's findings revealed significant insights into the performance of the vision model trained on synthetic images generated through diffusion models: The vision model showed exceptional performance in accurately identifying cases of genital warts. It achieved an accuracy rate of 96%, underscoring its effectiveness in medical image classification. For HPV cases, the model demonstrated a high precision of 99% and a recall of 94%. In normal cases, the precision was 95%, with an impressive recall of 99%. These metrics indicate the model's capability to correctly identify true positive cases and minimize false positives. The model achieved an F1-Score of 96% for HPV cases and 97% for normal cases. The high F1-Score across both categories highlights the balanced nature of the model's precision and recall, ensuring reliability and robustness in its predictions. We believe that by using the SynthVision methodology we propose, we will be able to develop accurate computer vision models requiring minimal data input for future medical emergencies.

# 1  Introduction

A significant issue faced today is in the development and training of computer vision models for detecting visual symptoms of emerging diseases, as it is critically hindered by the need for substantial, large datasets. This requirement for extensive data poses a significant challenge, particularly in emergent medical conditions like new natural endemics, pandemics or events of bioterrorism, where relevant datasets are inherently limited. Traditional data collection and aggregation methods are time-consuming and often insufficient to meet the rapid pace of new disease emergence and detection. Therefore, any delay in building detection models will result in a significant disease spread and burden on healthcare systems. This paper proposes an innovative solution to this predicament, SynthVision, by employing diffusion models for image generation, specifically focusing on creating synthetic medical images from a minimal set of guide images.

In the past, methods for the generation of synthetic images involved the use of Generative Adversarial Networks (GANs). The study by Aljohani & Alharbe (2022)[1] utilized Deep Pix2Pix GAN to create synthetic medical images and highlighted the potential of GANs in producing realistic and clinically accurate images, an approach that aligns with the ethos of our SynthVision project. However, GANs are frequently criticized for their lack of output diversity and unstable training [2]. Another limitation is the requirement of large datasets, which, as we have shown, are not readily available, especially in relation to medical emergencies.

Diffusion probabilistic models, a recent advancement in computer vision, have shown remarkable success in generating high-quality images. Khader et al. (2022)[3] demonstrated that these models can synthesize high-quality medical imaging data, such as MRI and CT images, and improve breast segmentation models' performance when data is scarce [1]. This is particularly relevant in the context of new diseases, where data scarcity is a significant barrier. The synthetic images generated through diffusion models are diverse and maintain high fidelity, ensuring their utility in training robust vision models.

The recent advancements in Denoising Diffusion Probabilistic Models (DDPMs), as exemplified by Nichol and Dhariwal's work [4], mark a significant leap in diffusion model capabilities. Their work demonstrated the process of image generation, providing high-quality results with optimized computational resources. The ability to reverse the diffusion process effectively, transforming noise into coherent imagery, aligns with our project's goal of harnessing minimal input for maximal output in computer vision models.

Furthermore, Ceritli et al. (2023)[5] explored the potential of diffusion models in generating realistic mixed-type Electronic Health Records (EHRs), demonstrating their advantage in generating more realistic synthetic data [5]. This research underscores the versatility of diffusion models in handling different data modalities, which is crucial in the context of healthcare, where data variety is paramount. The ability to generate realistic images rapidly and efficiently is crucial in responding to emerging healthcare challenges, such as the COVID-19 pandemic.

In summary, this paper aims to demonstrate the effectiveness of using diffusion models for synthetic image generation in healthcare, emphasizing their role in overcoming the challenges posed by sparse data availability. The approach can enrich the dataset needed for training effective vision models, thereby enhancing diagnostic accuracy and patient care in the face of rapidly evolving medical challenges.

# 2   Methods

## 2.1   Phase 1 - Fine-Tuning of Diffusion Models and Synthetic Image Generation

**Methodology for Generating Images of HPV Genital Warts Using Personalized Text-to-Image Diffusion Models**

Our approach utilizes a 4-step fine-tuning process of a text-to-image diffusion model, tailored specifically for generating images of genital warts in men.

To improve the fidelity of generated images representing Human Papillomavirus (HPV) genital warts, we utilized the DreamBooth technique to refine the capabilities of the Stable Diffusion 1.5 model. DreamBooth represents an advanced training methodology that enhances the entire diffusion model by using a minimal set of images to capture a particular subject or style effectively. This approach enables the production of precise and detailed visual content, closely mirroring the distinct attributes of HPV genital warts. The primary aim of this methodology is to produce images that are both high in fidelity and medically accurate, serving as invaluable resources for research, diagnostic processes, and educational initiatives within the medical community.

1. **Initial Image Collection and Model Personalization**:
    - In the personalization phase, we curated 10 clinically validated images of HPV Genital warts. These images represented the entire disease spectrum of HPV, ensuring all manifestations were represented. These images were uniformly formatted and paired with descriptive text files. The naming convention of the text files mirrored the corresponding pictures to streamline the training process.
    - This process involves creating a unique identifier for the subject class (e.g., "genital warts") and pairing these images with text prompts that include the unique identifier and the class name. The prompt is structured in a way to describe in clinical detail what is seen, there for instead of a using, "HPV warts on penis" as a prompt we use "Many early stage grey colored small warts around the mid shaft of the penis."
    - Concurrently, we apply a class-specific prior preservation loss. This technique leverages the model's semantic knowledge about the class ('genital warts'), promoting the generation of diverse yet class-specific images. This is achieved by incorporating the class name into the text prompts, enhancing the model's focus on generating images representative of genital warts.

2. **Super-Resolution Fine-Tuning**:
    - The second phase involves refining the super-resolution component of the model. Here, we use pairs of low-resolution and high-resolution images from our initial dataset.
    - This step is critical for preserving the intricate details of genital warts, ensuring that the generated images maintain fidelity to subtle clinical features, which are

imperative for accurate diagnosis and educational purposes. Fine tuning process illustrated in figure 1.

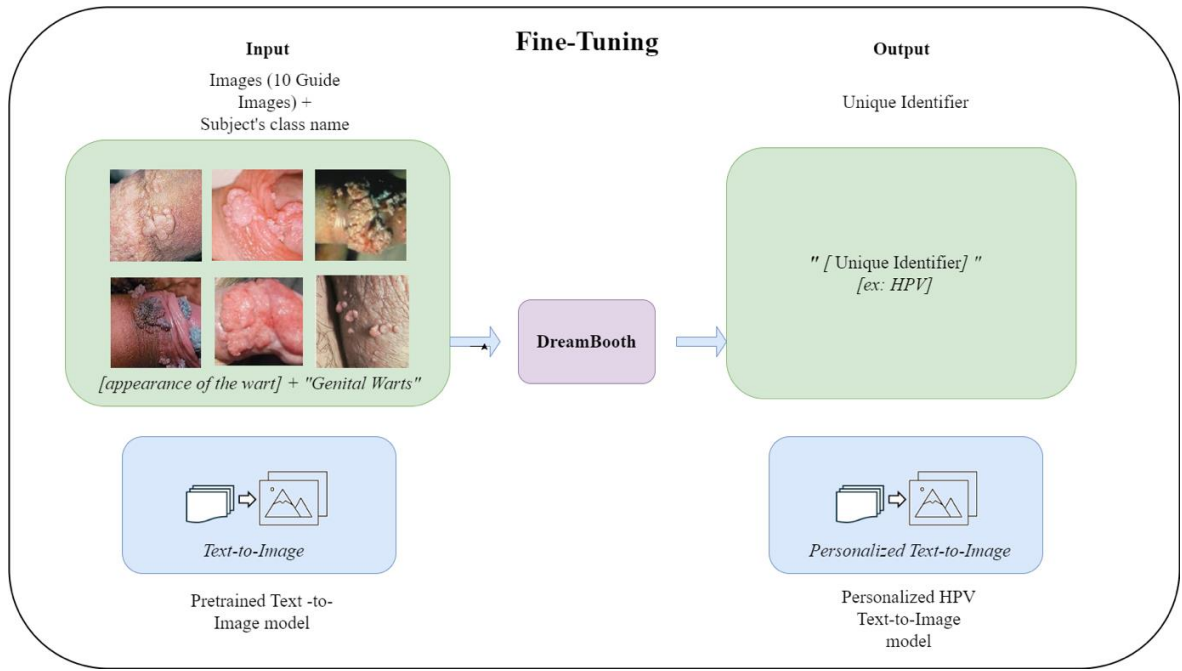

*Figure 1: Fine Tuning Process*

3. **Clinical Guidance and Model Adjustment in Initial Trials**:

   - Our initial trials with this method revealed challenges in rendering the genital area accurately. To address this, we strategically shifted the model's focus to primarily the warts, excluding the broader genital area from the image generation process.

   - This modification significantly enhanced the model's ability to produce clinically relevant and accurate representations of genital warts, minimizing distractions and inaccuracies related to the rendering of the genital area.

4. **Personalized HPV Text –to-image model training**

   The training was executed on a Tesla P100 GPU with 16GB of memory. Key training parameters included:

   - UNet Training Steps: Set at 2000, determined as optimal for the dataset's size and medical imagery nature.

   - UNet Learning Rate: Fixed at 2e-6, optimized for the dataset's small size and medical image processing complexity.

   - Text Encoder Training Steps: Established at 350 to balance adequate learning and overfitting prevention.

   - Text Encoder Learning Rate: Maintained at 4e-7 to minimize overfitting risks.

   - Image Resolution: Standardized at 512 for uniform image size and quality.

- Checkpoint Frequency: Set at every 500 steps for regular model monitoring and adjustments.

5. **Synthetic Image Generation and Clinical Validation**

    We generated a total of 630 synthetic images using carefully crafted prompts, generating 30 to 50 variants per prompt.

    - A trained physician then analyzed all 630 synthetic images to select images which were clinically accurate and representative of all the possible variations and severity levels of HPV genital warts in men.

    - From the images scrutinized, 130 images were found to be clinically inaccurate and not a realistic representation of HPV genital warts in men and thus excluded, leaving 500 synthetic images to be used in the final training dataset. Clinical Validation is illustrated in figure 2.

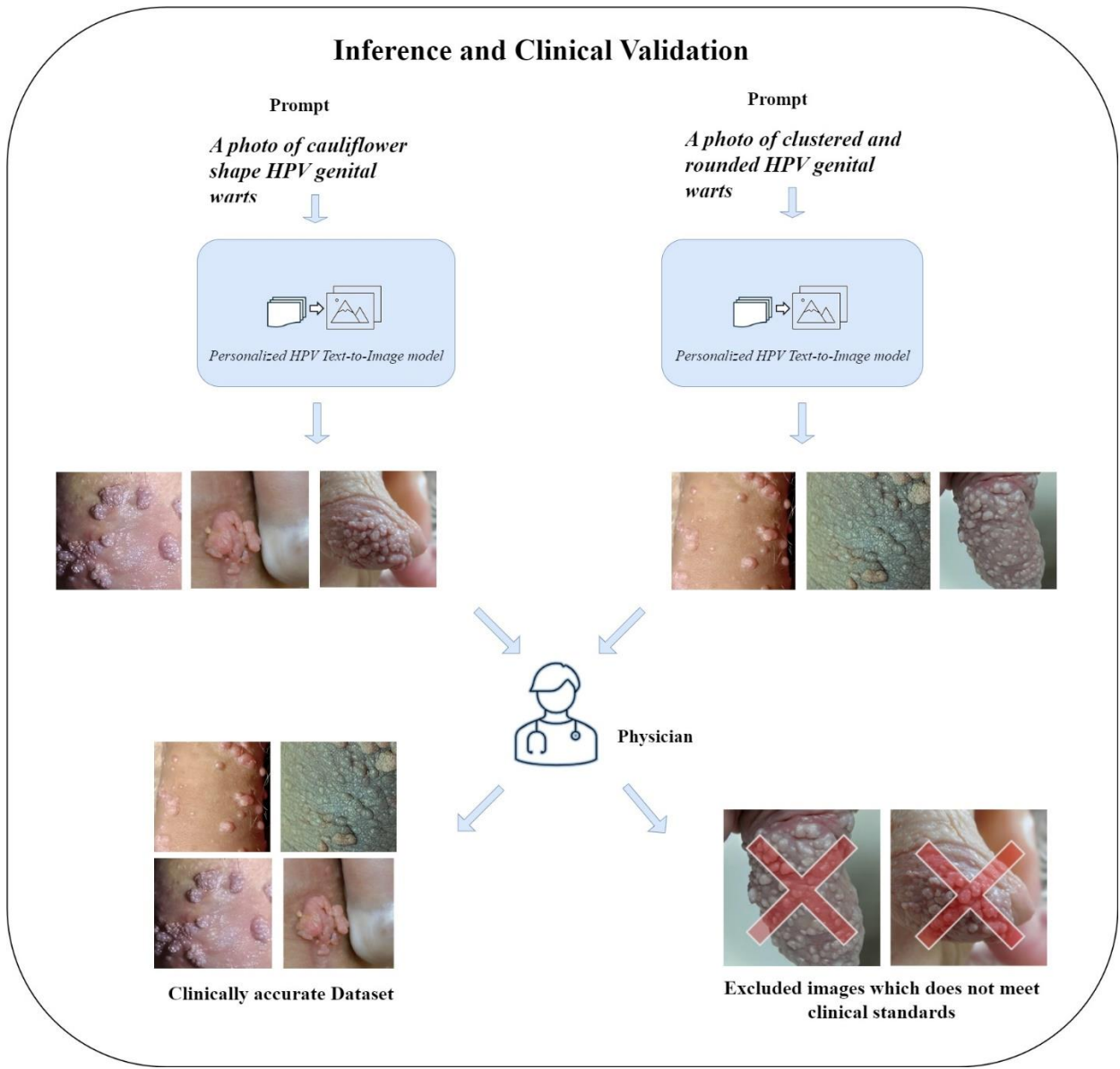

*Figure 2: Inference and Clinical Validation*

This methodological approach represents a significant advancement in the use of AI for medical image generation, offering a reliable and efficient means of producing clinically accurate images for a variety of applications in the healthcare sector.

## 2.2 Phase 2 – Computer Vision model Development

1. **Classification Model Development**

   The primary objective of the classification model was to showcase the effectiveness of synthetic data and diffusion models in generating diverse datasets. The aim was to develop a diagnostic model for HPV that can differentiate between HPV genital warts and normal cases (disease negative) using a training dataset exclusively composed of synthetic data.

2. **Dataset Preparation for Classification Model**

   As stated above, 500 synthetic images were clinically validated and selected for use, while the remaining 130 were discarded due to inadequate clinical relevance. The experiment utilized two classes across three distinct datasets: training, validation, and testing sets. The training set comprised 500 synthetic but clinically validated HPV genital wart images alongside 500 real images of a normal penis, using the synthetic images to represent the HPV class entirely. This set is fundamental for the initial model training, focusing on developing the model's predictive capabilities.

   The Testing dataset was curated by a trained physician to ensure that all possible variations and severity levels were represented. This was done to accurately gauge the performance of the computer vision model in detecting HPV warts of all possible variations.

   For model tuning and architectural decisions, a validation set consisting of 50 real and clinically verified images of normal penises and HPV genital warts was employed. This dataset ensures the model's performance is evaluated on unseen data, preventing overfitting and enhancing its generalization ability.

   Finally, the model's performance was assessed using a testing set, which included 70 clinically validated real images for each class, never introduced during the training phase. This set serves as an unbiased performance evaluator, simulating real-world scenarios to gauge the model's expected real-world efficacy.

| Data | Class | Description |
|---|---|---|
| Training Data | HPV | 630 HPV Genital warts images were generated from the diffusion model and scrutinized by a physician for accuracy, out of which 500 images were selected. |
| | Normal | 500 real images of a normal penis (Disease negative) |
| Validation Data | HPV | 50 real images of HPV Genital Warts |
| | Normal | 50 real images of a normal penis (Disease negative) |
| Test Data | HPV | 70 real images of HPV Genital Warts |
| | Normal | 70 real images of a normal penis (Disease negative) |

*Table 1: Dataset*

3. **Model Architecture**

The architecture employed is the ""ViT-Base-Patch16-224"", which pertains to the Vision Transformer configured for medium-scale models. It processes images in 16x16 pixel patches and is optimized for inputs of size 224x224 pixels. We have enhanced the original architecture by integrating attention dropout mechanisms to refine its focus during training.

4. **Model Parameters**

During the development of our model, we engaged in a comprehensive hyperparameter tuning process to optimize its performance. This process involved experimenting with various settings to identify the combination that yields the best accuracy. The key steps in our tuning process included:

**Epochs Adjustment:** We started by experimenting with the number of epochs, which is the number of times the learning algorithm works through the entire training dataset. Our experiments ranged from 20 to 200 epochs. Through these trials, we discovered that setting the number of epochs to 150 resulted in the highest accuracy for our model.

**Learning Rate Optimization:** Next, we focused on optimizing the learning rate, a critical parameter that affects the speed and quality of the learning process. We tested multiple learning rates, from 1e-2 down to 1e-5. The optimal learning rate was found to be 1e-4, as it provided the best balance between learning speed and model accuracy.

**Optimizer Selection:** Finally, we evaluated different optimizers, which are algorithms or methods used to change the attributes of the neural network such as weights and learning rate to reduce the losses. We experimented with Adam and RMSprop optimizers. Our findings indicated that RMSprop was the most effective optimizer for our model, leading to the highest accuracy.

Based on these experiments, we established the final settings for our model as follows:

- IMAGE_SIZE = 224x224 pixels
- BATCH_SIZE = 64
- EPOCHS = 150
- LEARNING_RATE = 1e-4
- Optimizer : RmsProp

# 3 Results

## 3.1 Confusion Matrix

The confusion matrix (Illustrated in figure 3) reveals that the model has a strong ability to distinguish between HPV-positive and normal cases. Specifically:

- Out of 70 HPV-positive cases, the model correctly identified 66, resulting in a high True Positive rate.
- It exhibited a low False Negative rate, with only 4 HPV-positive cases misclassified as normal.
- Impressively, the model demonstrated perfect specificity, with no False Positives; it did not misclassify any normal cases as HPV-positive.
- True Negatives were also maximized, with all 70 normal cases accurately identified.

The normalized figures within the confusion matrix underscore the model's proficiency: 94% of HPV cases were true positives. In comparison, 100% of normal cases were true negatives, indicating a solid balance between sensitivity and specificity.

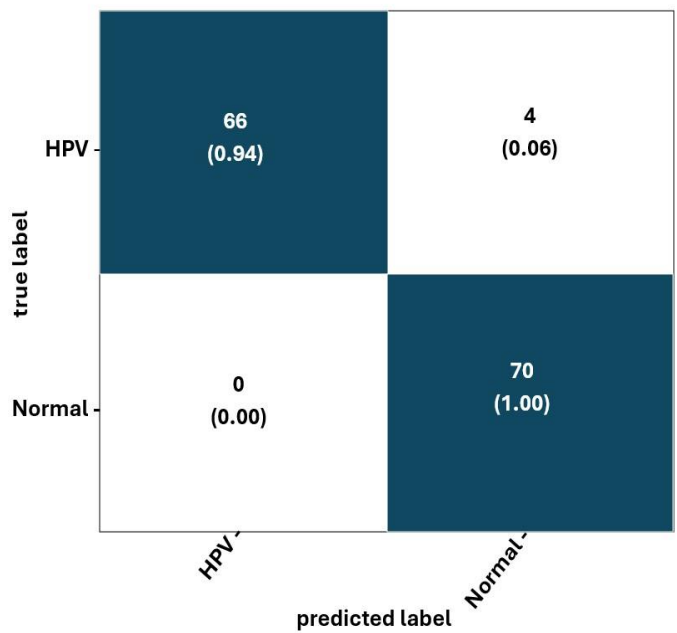

*Figure 3 : Confusion matrix*

## 3.2 Classification Report Analysis:

The classification report (Illustrated in figure 4) complements the confusion matrix with additional performance metrics:

- **Precision:** For HPV detection, the model achieved perfect precision (1.00), indicating there were no normal cases incorrectly labelled as HPV-positive. The precision for normal case detection was 0.95, showing a very small proportion of HPV cases incorrectly labelled as normal.
- **Recall:** The recall for HPV was 0.94, meaning that the model successfully identified 94% of all actual HPV cases. For normal cases, the recall was perfect (1.00), with all normal cases being correctly identified.
- **F1-Score:** The model attained an F1-score of 0.97 for both HPV and normal classifications, reflecting a robust balance between precision and recall, which is crucial for medical diagnostic models.
- **Support:** The dataset contained an equal number of HPV and normal cases (70 each), ensuring a balanced evaluation of the model's performance.

The model's overall accuracy was 97%, indicating exceptional performance across both classes of HPV and normal. The macro and weighted averages for precision, recall, and the F1-score were all at 0.97, confirming the model's consistent performance across different evaluation metrics.

```
Classification Report:
              precision    recall  f1-score   support

         HPV       1.00      0.94      0.97        70
      Normal       0.95      1.00      0.97        70

    accuracy                           0.97       140
   macro avg       0.97      0.97      0.97       140
weighted avg       0.97      0.97      0.97       140
```

*Figure 4 : Classification Report*

## 3.3 Receiver Operating Characteristic (ROC) Curve Analysis

The Receiver Operating Characteristic (ROC) curve (illustrated in figure 5) is a pivotal tool for assessing the true positive rate against the false positive rate at various threshold settings. Our model achieved an exemplary Area Under the Curve (AUC) of 0.993, which indicates outstanding discriminative power. The model's ROC curve closely approaches the upper left corner of the plot, suggesting a high true positive rate (sensitivity) coupled with a low false positive rate (1-specificity).

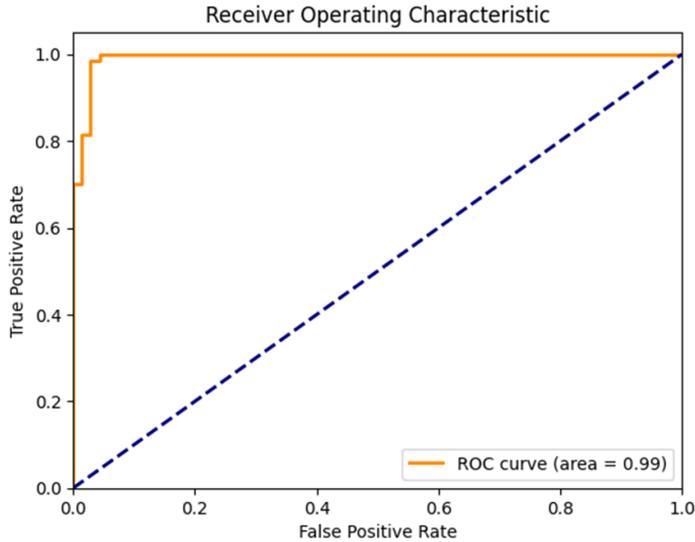

*Figure 5 : ROC Curve*

### 3.4  Overall Evaluation

The model exhibited high-performance metrics, indicating its potent capability to classify HPV and Normal cases with high accuracy, specificity, and sensitivity. The harmony between precision and recall, encapsulated by the F1-score, along with the high NPV and specificity, underscores the model's potential clinical utility.

The Average Precision score complements the AUC, reinforcing the model's proficiency in maintaining high precision across varying thresholds. This robustness is critical in clinical settings where the cost of false positives and negatives is high.

In conclusion, the presented model not only showcases high accuracy but also maintains excellent specificity and sensitivity, which are pivotal for a reliable diagnostic tool. These results reflect the model's potential for deployment in clinical decision-making processes, subject to further validation on independent datasets and real-world clinical environments.

## 4  Discussion

In this study, we attempted to prove the hypothesis that by using minimal amounts of real images, we can create a large synthetic dataset and use it to train a disease detection model. This model demonstrates performance that is comparable to or better than models built entirely on large datasets of real data. Our results show we have proven this hypothesis. This technique is extremely important in the context of rapidly emerging medical emergencies. In such situations, the availability of large and diverse datasets is often a significant bottleneck, particularly when faced with new pathogens or conditions. Generating accurate and representative synthetic images from limited data can significantly speed up the development and deployment of detection tools.

This study has successfully demonstrated the use of the DreamBooth training method and super-resolution fine-tuning to produce high-quality images of genital warts. The model's high precision, recall, and F1-Score values indicate its effectiveness in medical image classification. This achievement is similar to the advancements in the use of diffusion models for medical imaging, as described by Kazerouni et al. in 2022[6].

This breakthrough represents a significant milestone in medical image synthesis and disease detection model training using complete synthetic data, particularly in the fields of dermatology and sexually transmitted diseases.

## 4.1 Rapid Model Deployment

In the face of a new epidemic, time is of the essence. Traditional methods of collecting and annotating medical images can be time-consuming, delaying the development of necessary diagnostic tools. Our approach circumvents this by using latent diffusion models to create a synthetic dataset that is both representative of the condition and sufficiently diverse. This rapid generation of training data can lead to quicker development and deployment of vision models, which is crucial in the early stages of medical emergencies.

## 4.2 Adaptability to New Conditions

Epidemics often bring forth new or rare medical conditions for which existing datasets may be inadequate. The flexibility of our methodology allows for quick adaptation to new conditions. By fine-tuning with even a small number of images of a new condition, the model can learn to generate relevant synthetic images, aiding in the swift development of diagnostic models.

## 4.3 Case Study - COVID-19 Pandemic

A pertinent example is the COVID-19 pandemic. In the early stages, there was a dire need for rapid diagnostic tools to detect the virus in medical imaging, such as lung Chest X-rays(CXR) and CT scans. Using our approach, a model could have been quickly trained with a limited dataset of COVID-19 positive CXR images to generate additional synthetic images, accelerating the development of AI-based diagnostic tools.

## 4.5 Technical Achievements

The DreamBooth training method was crucial in this research, enabling the generation of subject-specific, high-quality images. The super-resolution fine-tuning process preserved intricate medical image details, critical for clinically relevant image synthesis, paralleling the focus on detail and quality in Chambon et al. (2022)'s study on adapting pre-trained models for medical imaging (Chambon, Bluethgen, Langlotz, & Chaudhari, 2022) [7].

## 4.6 Methodological Strengths and Limitations

While our approach innovatively addresses data scarcity and image quality, its dataset limitations could affect generalizability, a concern also noted by Ning et al. (2021)[8] in their study on diffusion skewness tensor imaging (Ning, Szczepankiewicz, Nilsson, Rathi, & Westin, 2021)[8].

## 4.7 Future Directions

This technology's extension to other medical conditions, particularly in data-limited scenarios, is promising. The ability to build new disease detection models with a small number of actual cases, along with the potential to apply these models in other image-based areas such as CXR, CT, MRI, and ultrasound images, is very promising and needs to be explored. Further research could improve model efficiency and reduce computational costs, inspired by Tanno et al. (2021)'s[9] work in uncertainty modelling in diffusion MRI (Tanno, Worrall, Kaden, & Alexander, 2021)[9].

In conclusion, the ability to use minimal data to quickly generate large amount of synthetic images and train accurate computer vision models is a game-changer in the context of new medical emergencies and data limited scenarios. It enables rapid development and deployment of diagnostic models, a crucial factor in controlling and managing outbreaks. This approach addresses current challenges in medical imaging and positions us to be more proactive and responsive in future epidemic scenarios.